\documentclass[11pt,onecolumn]{article}

\usepackage{a4wide}
\usepackage[]{amssymb}
\usepackage[]{amsmath}
\usepackage{url}
\usepackage[ansinew]{inputenc}
\usepackage{graphicx,color}

\newcommand{\rb}[1]{\raisebox{1.5ex}[-1.5ex]{#1}}

\begin{document}

\title{Curved Gabor Filters\\for Fingerprint Image Enhancement}
\date{}
\author{Carsten Gottschlich
\thanks{Carsten Gottschlich is with the Institute for Mathematical Stochastics, 
University of G\"ottingen, Goldschmidtstr. 7, 37077 G\"ottingen, Germany. 
Phone: +49-(0)551-39172126. 
Fax: +49-(0)551-3913505.
Email: gottschlich@math.uni-goettingen.de}}
\maketitle

\section*{Abstract}

Gabor filters play an important role
in many application areas
for the enhancement of various types of images
and the extraction of Gabor features.
For the purpose of enhancing curved structures in noisy images,
we introduce curved Gabor filters\footnote{This paper is an arXiv preprint version of \cite{Gottschlich2012}.}
which locally adapt their shape to the direction of flow.
These curved Gabor filters enable the choice of filter parameters 
which increase the smoothing power without creating artifacts in the enhanced image.
In this paper, curved Gabor filters are applied
to the curved ridge and valley structure
of low-quality fingerprint images.
First, we combine two orientation field estimation methods
in order to obtain a more robust estimation for very noisy images.
Next, curved regions are constructed by following the respective local orientation
and they are used for estimating the local ridge frequency.
Lastly, curved Gabor filters are defined based on curved regions and
they are applied for the enhancement of low-quality fingerprint images.
Experimental results on the FVC2004 databases show improvements of this approach
in comparison to state-of-the-art enhancement methods.

\section*{Index Terms} Curved Gabor filter, ridge frequency estimation, curved regions, curvature, FVC2004,
image enhancement, orientation field estimation, fingerprint recognition, verification tests, biometrics.

\section{Introduction and Related Work}

Gabor functions \cite{Gabor1946},
in the form of Gabor filters (GFs) \cite{Gabor1965} and Gabor wavelets \cite{Lee1996},
are applied for a multitude of purposes 
in many areas of image processing and pattern recognition.
Basically, the intentions for using
GF and log-GF \cite{Field1987} 
can be grouped into two categories:
first, GF aim at enhancing images \cite{LindenbaumFischerBruckstein1994} 
and the second common goal is to  
extract Gabor features obtained from responses of filterbanks.
Typical fields of application include:

\clearpage

\paragraph{Texture}

Texture segmentation \cite{DunnHiggins1995} 
and classification \cite{LiaoLawChung2009},
with applications such as e.g. recognizing species of tropical wood \cite{YusofRosliKhalid2009}
or classifying developmental stages of fruit flies \cite{ZhongChenZhang2009}.

\paragraph{Medical and biological applications}

In medical imaging, GFs are applied for the enhancement 
of structures like e.g.
finger veins \cite{ZhangYang2009} and
muscle fibers in ultrasound images \cite{ZhouZheng2009},
for the detection of blood vessels in retinal images \cite{WuZhangLiuBauman2006},
as well as for many other tasks like e.g. 
analyzing event-related brain activity \cite{SinkkonenEtalii1995},
assessing osteoporosis in radiographs \cite{MengkoTjandraprmudito2002}
and for modeling the behavior of simple cells in the mammalian visual cortex \cite{Daugman1985}.

\paragraph{Optical character recognition}

GFs are utilized for text segmentation \cite{JainBhattacharjee1992},
character recognition \cite{RamanathanEtalii2009},
font recognition \cite{HaTianZhang2005},
and license plate recognition \cite{CanerGecimAlkar2008}.

\paragraph{Object recognition}

Objects can be detected by GFs \cite{JainRathaLakshmanan1996},
e.g. cars \cite{LimGuntoro2002}.
Moreover, GFs can be used for performing content-based image retrieval \cite{Barbu2009}.

\paragraph{Biometrics}

Gabor functions play an important role in biometric recognition.
They are employed for many physical or behavioral traits including
iris \cite{Daugman2004},
face \cite{DucFischerBigun1999}, 
facial expression \cite{Rose2006}, 
speaker \cite{MildnerEtAlii2007},
speech \cite{WuZhangShi2009},
emotion recognition in speech \cite{HeLechMaddageAllen2009},
gait \cite{TaoLiWuMaybank2007},
handwriting \cite{ZhuTanWang2000},
palmprint \cite{ZhangKongYouWong2003},
and fingerprint recognition.

\paragraph{Fingerprint recognition}

Gabor filterbanks are used for the
segmentation \cite{AlonsoFierrezOrtega2005}
and quality estimation \cite{ShenKotKoo2001} of fingerprint images,
for core point estimation \cite{LeeWang1999},
classification \cite{JainPrabhakarHong1999}
and fingerprint matching based on Gabor features \cite{JainPrabhakarHongPankanti2000,LeeWang1999}.
GFs are also employed for generating synthetic fingerprints \cite{CappelliErolMaioMaltoni2000}.
The use of GF for fingerprint image enhancement was introduced in \cite{HongWanJain1998}.

\begin{figure}
  \begin{center}
    \includegraphics[width=0.5\textwidth]{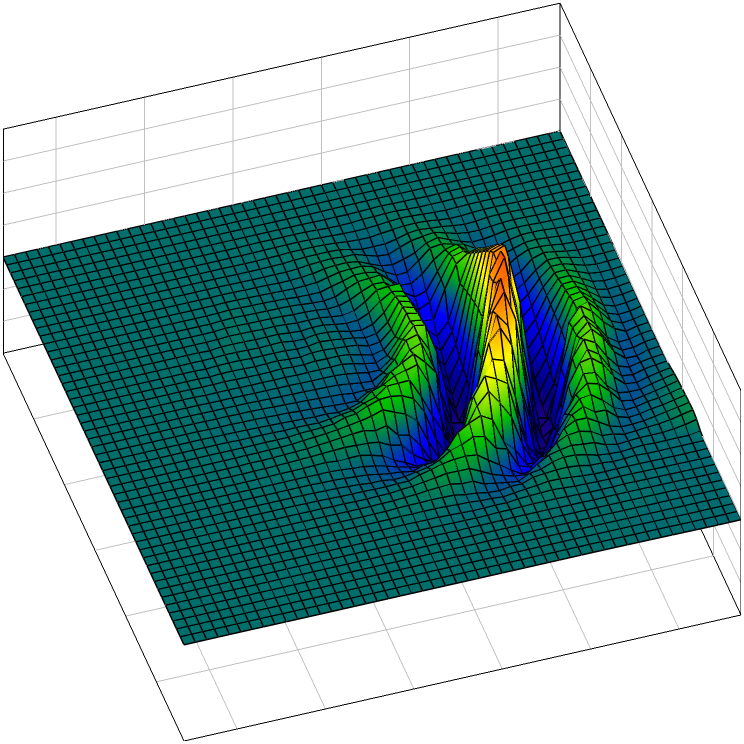}
  \end{center}
  \caption{Visualization of a curved Gabor filter for a pixel near a core point. 
           Filter parameters are: $\sigma_{x} = 6.0$, $\sigma_{y} = 8.0$.
           The underlying curved region consists of 33 parallel curved lines
           with 65 points for each line.
           \label{figCurvedGabor3D}}
\end{figure}

All aforementioned applications have in common that
they use \textit{straight} Gabor filters,
i.e. the x- and y-axis of the window underlying the GF are straight lines which are orthogonal.
Having the natural curvature inherent to fingerprints in mind,
we propose \textit{curved Gabor filters} (see Figure \ref{figCurvedGabor3D}) for image enhancement \cite{Gottschlich2010PhD}.
A GF can be regarded as an anisotropic diffusion filter
which smooths along the orientation 
and performs inverse diffusion orthogonal to the orientation \cite{SpiraSocheneKimmel2005}.
The basic idea is to adopt the GF to the curved structure 
and smooth along the bent ridges and valleys.
While this paper focuses on fingerprint image enhancement, 
curved Gabor filters might also be useful 
in other fields of application, 
e.g. for the enhancement of curved structures 
like muscle fibers, cell filaments or annual rings in tree discs.

\subsection{Fingerprint Image Enhancement} 

Image quality \cite{AlonsoEtAlii2007} has a big impact on the performance 
of a fingerprint recognition system (see e.g. \cite{NFIQ} and \cite{ChenDassJain2005}). 
The goal of image enhancement is to improve the overall performance
by optimally preparing input images for later processing stages. 
Most systems extract minutiae from fingerprints \cite{HandbookFR2009},
and the presence of noise can interfere with the extraction. As a result,
true minutiae may be missed and false minutiae may be detected,
both having a negative effect on the recognition rate. 
In order to avoid these two types of errors, 
image enhancement aims at improving 
the clarity of the ridge and valley structure.
With special consideration to the typical types of noise occurring in fingerprints,
an image enhancement method should have three important properties:
\begin{itemize}

\item reconnect broken ridges, e.g. caused by dryness of the finger or scars;

\item separate falsely conglutinated ridges, e.g. caused by wetness of the finger or smudges;

\item preserve ridge endings and bifurcations.

\end{itemize}

\noindent
Enhancement 
of low quality images (occurring e.g. in all databases of FVC2004 \cite{FVC2004})
and very low quality prints like latents (e.g. NIST SD27 \cite{NistSD27}) 
is still a challenge. 
Techniques based on contextual filtering are widely used 
for fingerprint image enhancement \cite{HandbookFR2009} 
and a major difficulty lies in an automatic and reliable estimation of the local context,
i.e. the local orientation and ridge frequency as input of the GF.
Failure to correctly estimate the local context can lead to the creation
of artifacts in the enhanced image \cite{Jiang2001} which consequently tends to increase
the number of identification or verification errors.

For low quality images, there is a substantial risk that an image enhancement step
may impair the recognition performance 
as shown in \cite{FronthalerKollreiderBigun2008} 
(results are cited in Table \ref{tabBZ3} of Section \ref{SecResults}).
The situation is even worse for very low quality images, 
and current approaches focus on minimizing the efforts required by a human expert 
for manually marking information 
in images of latent prints (see \cite{CappelliMaioMaltoni2009} and \cite{YoonFengJain2010}).

The present work addresses these challenges as follows:
in the next section, two state-of-the-art methods for orientation field estimation 
are combined for obtaining an estimation which is more robust than each individual one.
In Section \ref{SecRF}, curved regions are introduced 
and employed for achieving a reliable ridge frequency estimation.
Based on the curved regions, 
in Section \ref{SecCGF} curved Gabor filters are defined.
In Section \ref{SecResults}, all previously described methods are combined
for the enhancement of low quality images from FVC2004
and performance improvements in comparison to existing methods are shown.
The paper concludes with a discussion of the advantages and drawbacks 
of this approach, as well as possible future directions in Section \ref{SecConclusions}.

\section{Orientation Field Estimation}
\label{SecOF}

In order to obtain a robust orientation field (OF) estimation for low quality images,
two estimation methods are combined:
the line sensor method \cite{GottschlichMihailescuMunk2009}
and the gradients based method \cite{BazenGerez2002} 
(with a smoothing window size of 33 pixels).
The OFs are compared at each pixel. 
If the angle between both estimations is smaller than a threshold (here $t = 15\,^\circ$~),
the orientation of the combined OF is set to the average of the two.
Otherwise, the pixel is marked as missing.
Afterwards, all inner gaps are reconstructed 
and up to a radius of 16 pixels, 
the orientation of the outer proximity is extrapolated,  
both as described in \cite{GottschlichMihailescuMunk2009}.

Results of verification tests on all 12 databases 
of FVC2000 to 2004 \cite{FVC2000,FVC2002,FVC2004} 
showed a better performance of the combined OF applied for contextual image enhancement 
than each individual OF estimation \cite{Gottschlich2010PhD}.
The OF being the only parameter that was changed, 
lower equal error rates can be interpreted as an indicator that the combined OF 
contains fewer estimation errors than each of the individual estimations.
Simultaneously, we regard the combined OF as a segmentation 
of the fingerprint image into foreground (endowed with an OF estimation) and background.

The information fusion strategy for obtaining the combined OF 
was inspired by \cite{PreddOshersonKulkarniPoor2008}.
The two OF estimation methods can be regarded as judges or experts 
and the orientation estimation for a certain pixel as a judgment.
If the angle between both estimations is greater than a threshold $t$,
the judgments are considered as incoherent,
and consequently not averaged. 
If an estimation method provides no estimation for a pixel,
it is regarded as abstaining. 
Orientation estimations for pixels with incoherent or abstaining judges
are reconstructed or extrapolated  
from pixels with coherent judgments.

\section{Ridge Frequency Estimation Using Curved Regions} \label{SecRF}

\begin{figure}
  \begin{center}
    \includegraphics[height=0.5\textwidth]{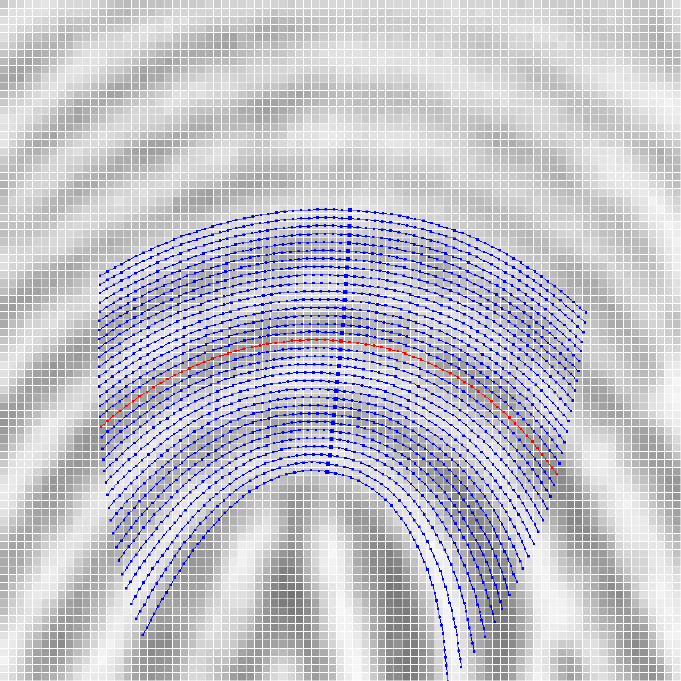}
    \hspace{1cm}
    \includegraphics[height=0.5\textwidth]{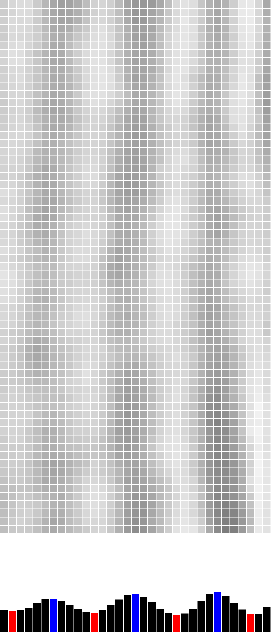}
  \end{center}
  \caption{A curved region (left) with
           33 lines (central line in red, 
           parallel lines start at the blue squares) and 65 points per line.
           The gray-level profile is obtained by averaging the interpolated gray values
           along each line, followed by the detection of
           local minima (red) and maxima (blue). 
           \label{figCurvedRegion1}}
\end{figure}

In \cite{HongWanJain1998}, a ridge frequency (RF) estimation method was proposed
which divides a fingerprint image into blocks of $16 \times 16$ pixels,
and for each block, it obtains an estimation from an oriented window of $32 \times 16$ pixels
by a method called `x-signature' which detects peaks in the gray-level profile.
Failures to estimate a RF, e.g. caused due to presence of noise, curvature or minutiae,
are handled by interpolation and outliers are removed by low-pass filtering.
In our experience, this method works well for good and medium quality prints,
but it encounters serious difficulties obtaining a useful estimation
when dealing with low quality prints.

In this section, we propose a RF estimation method 
following the same basic idea - to obtain an estimation from the gray-level profile -
but which bears several improvements in comparison to \cite{HongWanJain1998}:
(i)~the profile is derived from a curved region which is different in shape and size
from the oriented window of the x-signature method,
(ii) we introduce an information criterion (IC) for the reliability of an estimation and
(iii) depending on the IC, the gray-level profile is smoothed with a Gaussian kernel,
(iv) both, minima and maxima are taken into account and 
(v) the inverse median is applied for the RF estimate.

\begin{figure}
  \begin{center}
    \includegraphics[width=0.4\textwidth]{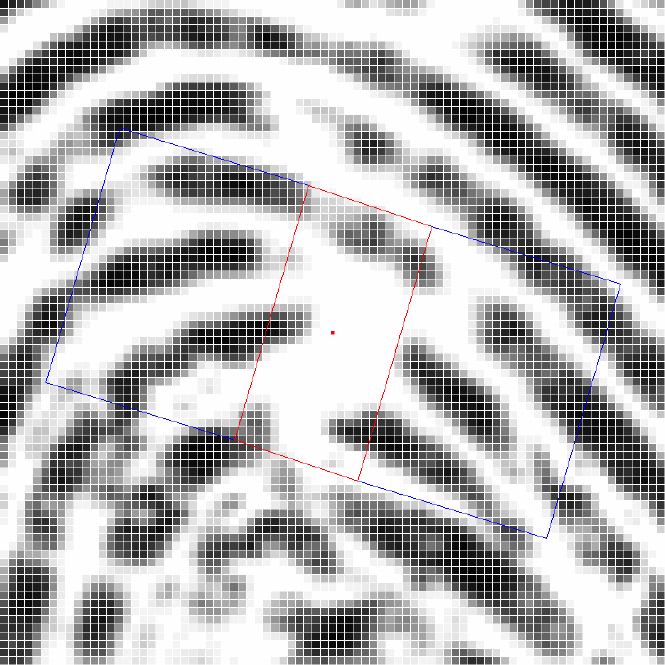}        
    \includegraphics[width=0.4\textwidth]{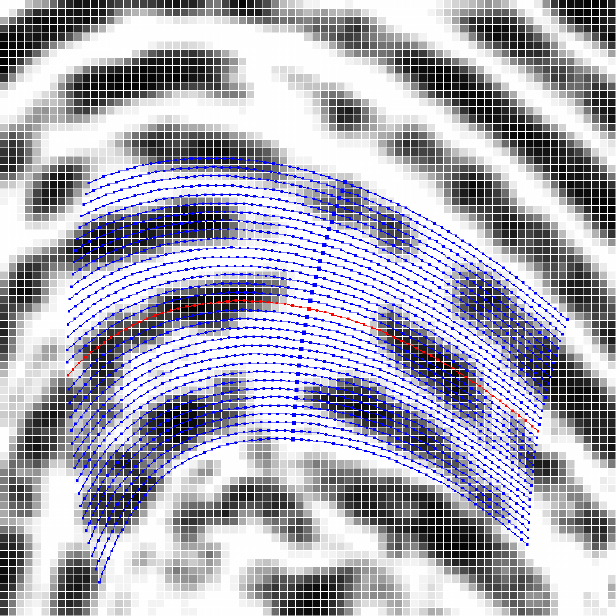}
  \end{center}
  \caption{Comparing an oriented window (red rectangle) 
           of $32 x 16$ pixels as used by the x-signature method (left) 
           and a curved region consisting of 33 parallel lines and 65 points per line.           
           Noise can cause the x-signature method to fail,
           because the oriented window may contain an insufficient amount
           of information. Magnifying the window (blue) along the local orientation does not remedy 
           these deficiencies in regions of curvature and would lead to an erroneous gray level profile.
           The RF estimation based on curved regions (right) overcomes these limitations by considering 
           the change of local orientation, i.e. curvature. 
           \label{figComparisonRF}}
\end{figure}

If the clarity of the ridge and valley structure is disturbed by noise, 
e.g. caused by dryness or wetness of the finger,
an oriented window of $32 \times 16$ pixels may not contain a sufficient amount of information 
for a RF estimation (e.g. see Figure \ref{figComparisonRF}, left image).
In regions where the ridges run almost parallel,
this may be compensated by averaging over larger distances along the lines.
However, if the ridges are curved, the enlargement of the rectangular window 
does not improve the consistency of the gray-profile,
because the straight lines cut neighboring ridges and valleys.
In order to overcome this limitation,
we propose \textit{curved regions} which adapt their shape to the local orientation.
It is important to take the curvature of ridges and valleys into account,
because about 94 \% of all fingerprints 
belong to the classes right loop, whorl, left loop and tented arch \cite{JainPrabhakarPankanti2002},
so that they contain core points and therefore regions of high curvature.

\subsection{Curved Regions} 

Let $(x_c, y_c)$ be the center of a curved region 
which consists of $2p + 1$ parallel curves and $2q + 1$ points along each curve.
The midpoints (depicted as blue squares in Figure \ref{figCurvedRegion1}) 
of the parallel curves are initialised by following both directions orthogonal 
to the orientation for $p$ steps of one pixel unit, starting from the central pixel $(x_c, y_c)$ (red square).
At each step, the direction is adjusted, so that it is orthogonal to the local orientation.
If the change between two consecutive local orientations is greater than a threshold,
the presence of a core point is assumed, and the iteration is stopped.
Since all x- and y-coordinates are decimal values, the local orientation is interpolated.
Nearest neighbour and bilinear interpolation using the orientation 
of the four neighboring pixels are examined in Section \ref{SecResults}.
Starting from each of the $2p + 1$ midpoints,
curves are obtained by following the respective local orientation 
and its opposite direction (local orientation $\theta + \pi$) 
for $q$ steps of one pixel unit, respectively.

\subsection{Curvature estimation} 

\begin{figure}
  \begin{center}
    \includegraphics[width=0.4\textwidth]{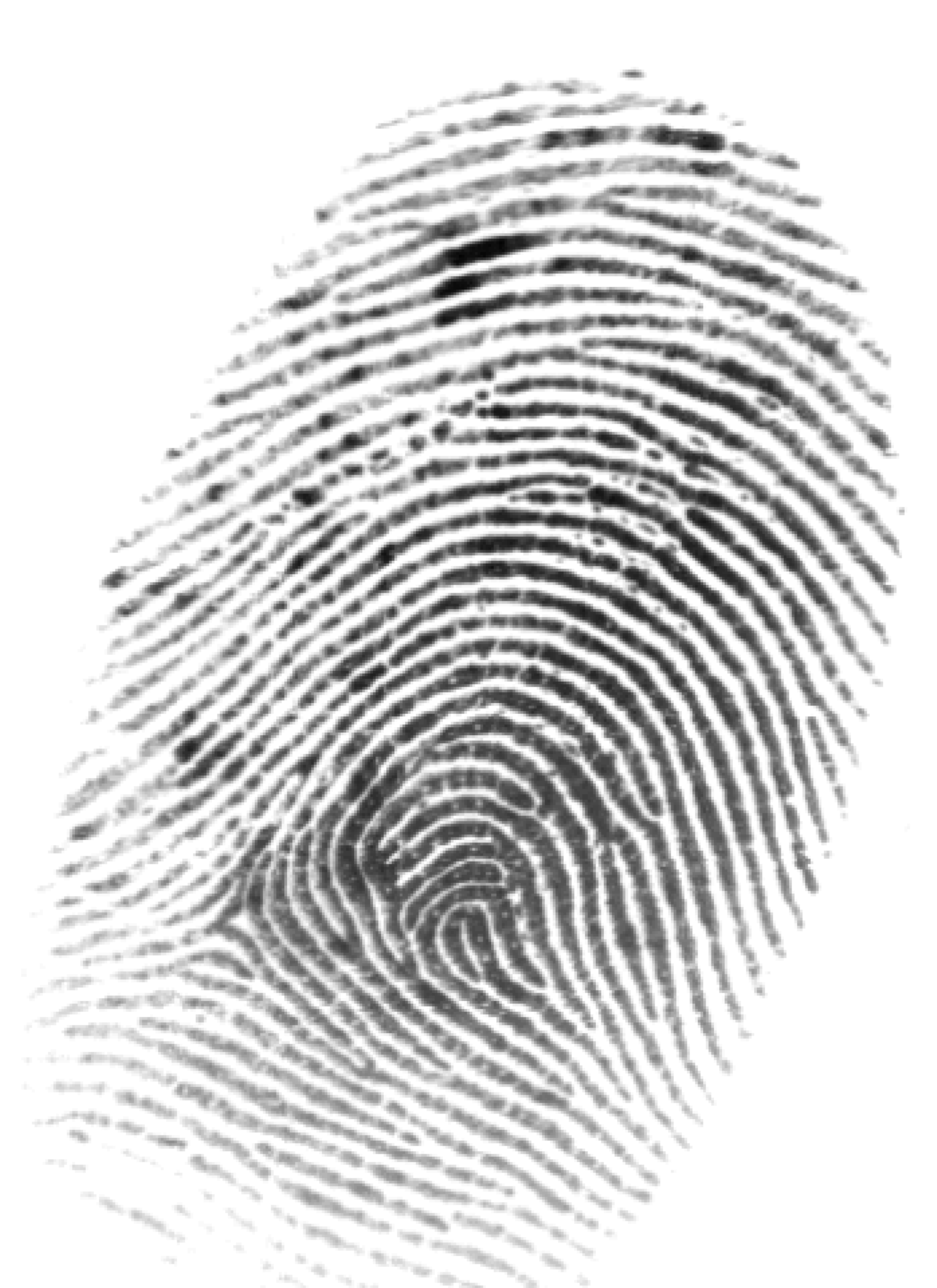}   
    \includegraphics[width=0.4\textwidth]{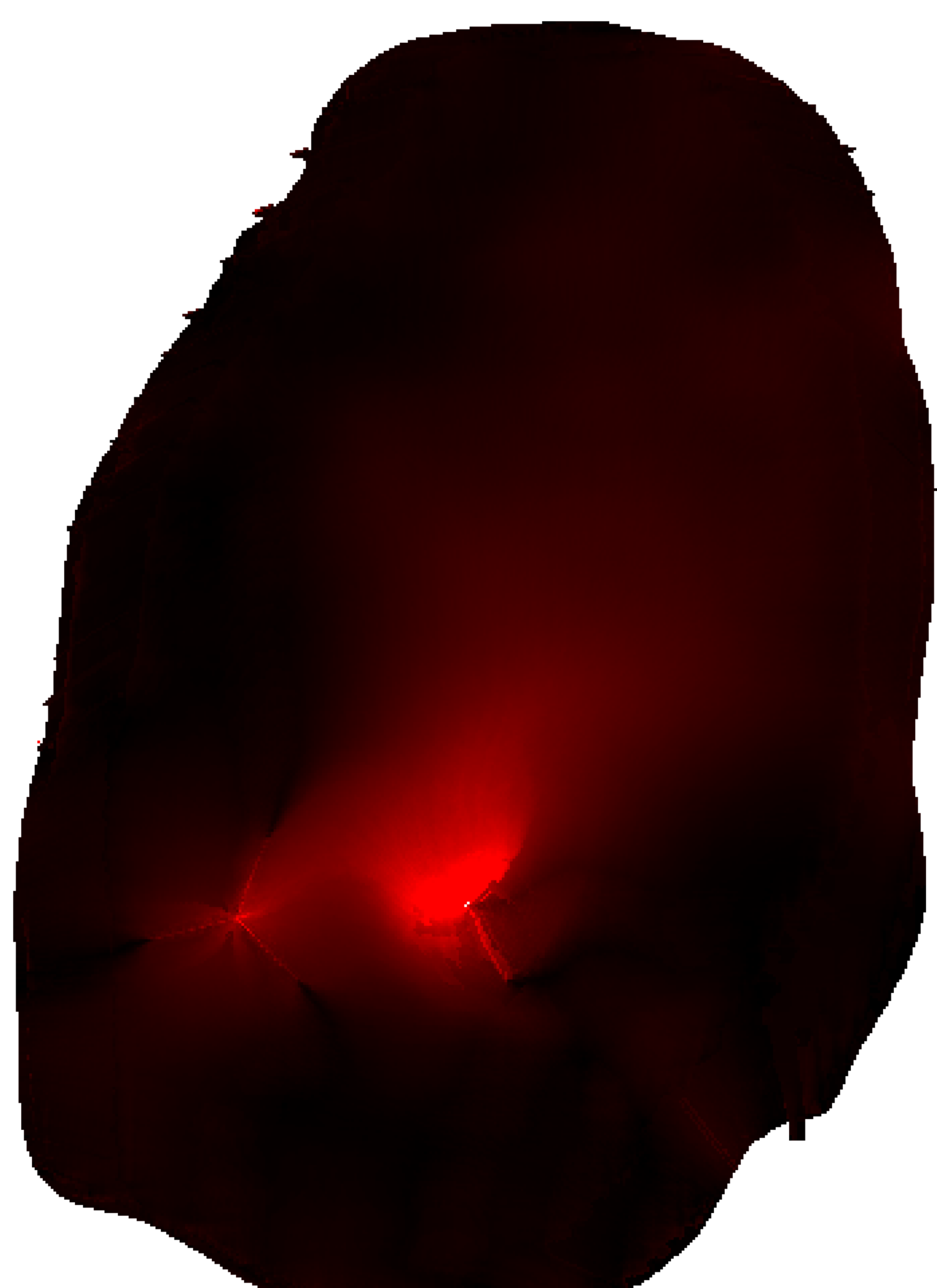}   
  \end{center}
  \caption{Points of high curvature are visualized by a high intensity of red. 
           The core and delta points are clearly discernible in the curvature image on the right.
           \label{figCurvature}}
\end{figure}

As a by-product of constructing curved regions,
a pixel-wise estimate of the local curvature is obtained 
using the central curve of each region (cf. the red curves in Figures 
\ref{figCurvedRegion1} and \ref{figComparisonRF}). 
The estimate is computed by adding up the absolute values of differences in orientation
between the central point of the curve and the two end points.
The outcome is an estimate of the curvature, 
i.e. integrated change in orientation along a curve (here: of 65 pixel steps).
For an illustration, see Figure \ref{figCurvature}.
The curvature estimate can be useful for singular point detection,
fingerprint alignment or as additional information at the matching stage.

\subsection{Ridge Frequency Estimation} 

\begin{figure}
  \begin{center}
    \includegraphics[width=0.25\textwidth]{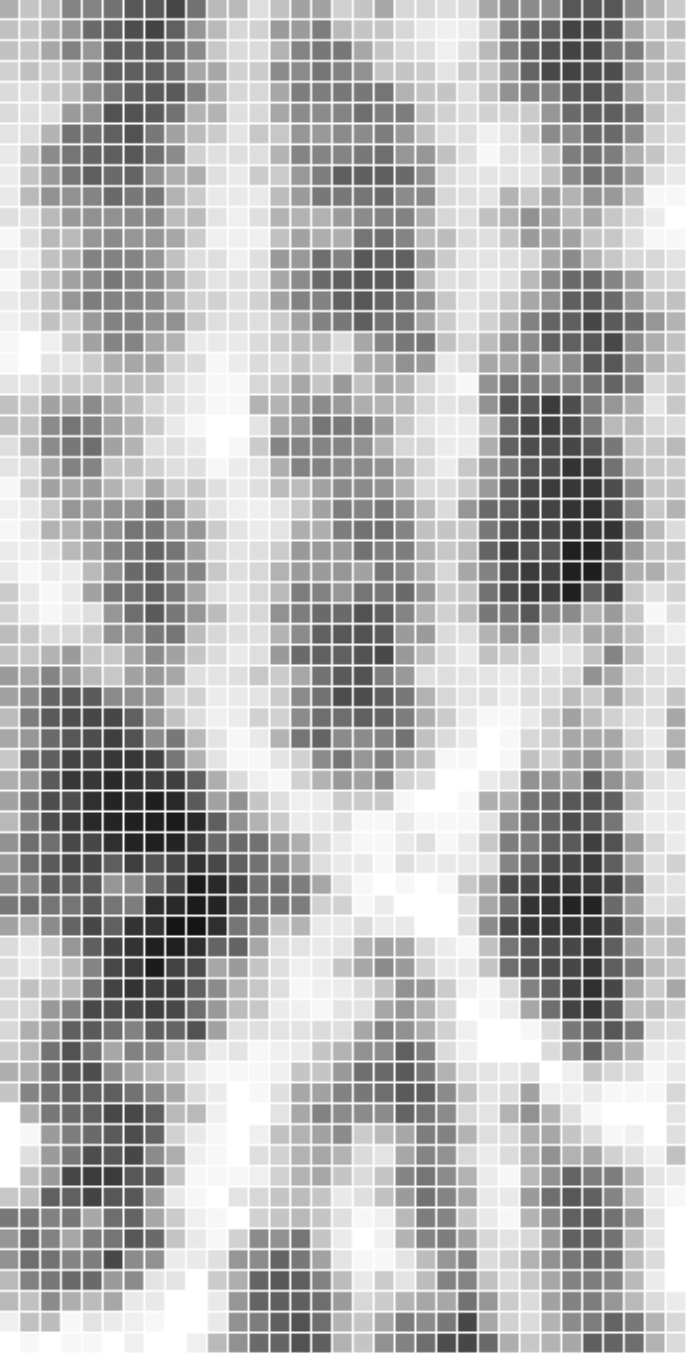} \\    
    \includegraphics[width=0.25\textwidth]{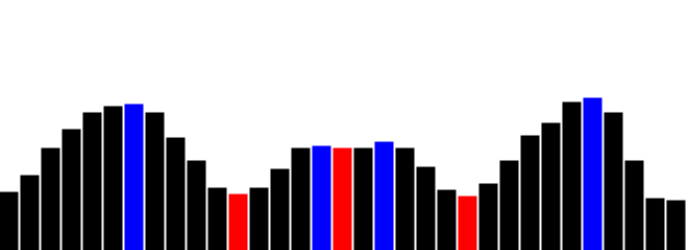} \\    
    \includegraphics[width=0.25\textwidth]{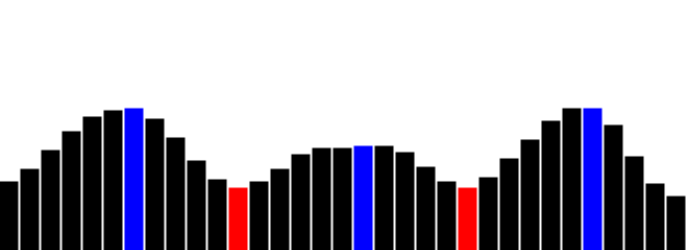}
  \end{center}
  \caption{The RF estimation for the depicted example is complicated by noise 
           and a bifurcation (in the bottom area of the top image).
           The profile obtained by averaging along the lines (center)
           contains false minima and maxima which results in the following
           inter-extrema distances (IEDs): 9, 3, 10 (between maxima), 5, 6 (between minima).           
           Median: 6 and proportion $p_{maxmin} = 10/3 = 3.3$.
           Here, false extrema are removed by one smoothing iteration 
           with a Gaussian kernel (size=7, $\sigma=1.0$).
           IEDs for the profile at the bottom: 11, 11, 11.
           Median: 11 and $p_{maxmin} = 1.0$.
           \label{figProfile1}}
\end{figure}

Gray values at the decimal coordinates of the curve points 
are interpolated. In this study, three interpolation methods are taken into account:
nearest neighbor, bilinear
and bicubic \cite{Russ2000} (considering 1, 4 and 16 neighboring pixels for the gray value interpolation, respectively).
The gray-level profile is produced by averaging 
the interpolated gray values along each curve (in our experiments,
the minimum number of valid points is set to 50\% of the points per line).
Next, local extrema are detected and the distances between consecutive minima and consecutive maxima are stored.

The RF estimate is the reciprocal of the median of the inter-extrema distances (IEDs).
The proportion $p_{maxmin}$ of the largest IED to the smallest IED is regarded 
as an information parameter for the reliability of the estimation:
\begin{displaymath}
p_{maxmin} = \frac{\displaystyle \max_{i \in IED} i}{\displaystyle \min_{i \in IED} i}
\end{displaymath}
Large values of $p_{maxmin}$ are considered as an indicator 
for the occurrence of false extrema in the profile (see Figure \ref{figProfile1}).
or for the absence of true extrema.
Only RF estimations where $p_{maxmin}$ is below a threshold 
are regarded as valid (for the tests in Section \ref{SecResults}, we used $thr_{p_{maxmin}} \leq 1.5$).
If $p_{maxmin}$ of the gray-level profile produced by averaging along the curves exceeds the threshold,
then, in some cases it is still possible to obtain a feasible RF estimation
by smoothing the profile which may remove false minima and maxima,
followed by a repetition of the estimation steps (see Figure \ref{figProfile1}).
A Gaussian with a size of 7 and $\sigma = 1.0$ was applied in our study,
and a maximum number of three smoothing iterations was performed.
In an additional constraint we require
that at least two minima and two maxima are detected and
the RF estimation is located within an appropriate range 
of valid values (between $\frac{1}{3}$ and $\frac{1}{25}$).
As a final step, the RF image is smoothed by averaging over a window of size $w = 49$ pixels.

\clearpage

\section{Curved Gabor Filters} \label{SecCGF}

\subsection{Definition} 

The Gabor filter is a two-dimensional filter formed 
by the combination of a cosine with a two-dimensional Gaussian function
and it has the general form:

\begin{equation}
{g}(x, y, \theta, f, \sigma_{x}, \sigma_{y}) = \exp \left\{ - \frac{1}{2} \left[ \frac{x_{\theta}^2}{\sigma_{x}^2} + \frac{y_{\theta}^2}{\sigma_{y}^2} \right] \right\} \cdot \cos \left( 2\pi \cdot f \cdot x_{\theta} \right)
\end{equation}

\begin{equation}
x_{\theta} = x \cdot \cos \theta + y \cdot \sin \theta 
\end{equation}

\begin{equation}
y_{\theta} = - x \cdot \sin \theta + y \cdot \cos \theta
\end{equation}

\noindent
In (1), the Gabor filter is centered at the origin. 
$\theta$ denotes the rotation of the filter related to the x-axis
and $f$ the local frequency.
$\sigma_{x}$ and $\sigma_{y}$ signify the standard deviation of the Gaussian function 
along the x- and y-axis, respectively.

A curved Gabor filter is computed by mapping
a curved region to a two-dimensional array,
followed by a point-wise multiplication with an unrotated GF ($\theta = 0$).
The curved region $C_{i,j}$ centered in $(i, j)$ 
consists of $2p + 1$ parallel lines and $2q + 1$ points along each line.
The corresponding array $A_{i,j}$ contains the interpolated gray values (see right image in Figure \ref{figCurvedRegion1}).
The enhanced pixel $E(i, j)$ is obtained by:

\begin{equation} \label{eqCGF}
{E(i, j, A_{i,j}, f_{(i,j)})} = \sum_{k=0}^{2p + 1} \  \sum_{l=0}^{2q + 1} A(k,l) \cdot g(k-p, l-q, 0, f_{(i,j)}, \sigma_{x}, \sigma_{y})
\end{equation}

Finally, differences in brightness are compensated by a locally adaptive normalization
(using the formula from \cite{HongWanJain1998} who proposed a global normalization 
as a first step before the OF and RF estimation, and the Gabor filtering).
In our experiments, the desired mean and standard deviation were set to 127.5 and 100, respectively,
and neighboring pixels within a circle of radius $r = 16$ were considered.

\subsection{Parameter Choice}

\begin{figure}
  \begin{center}
    \includegraphics[width=0.24\textwidth]{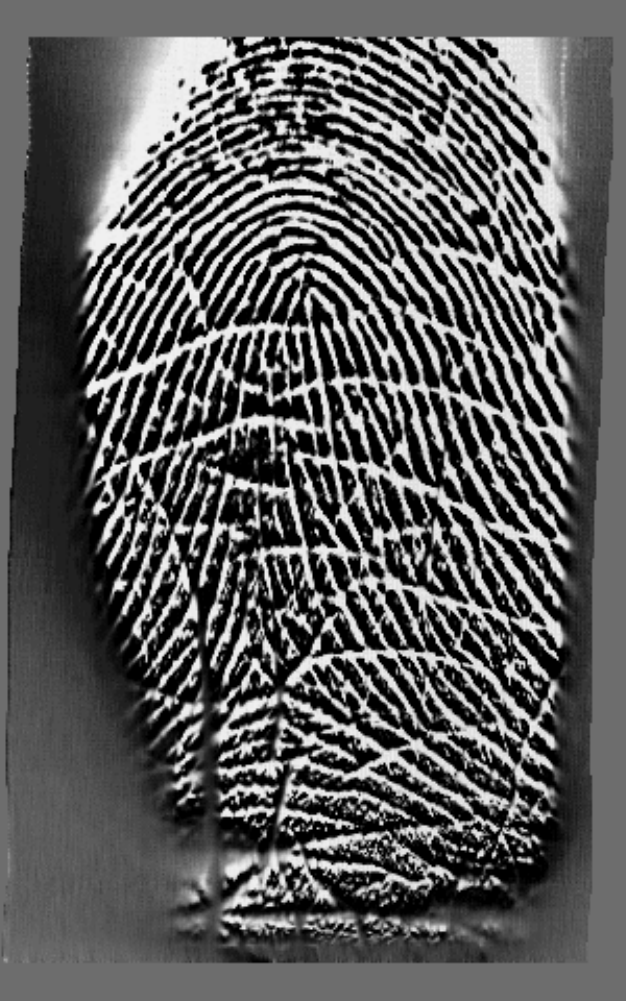}
    \includegraphics[width=0.24\textwidth]{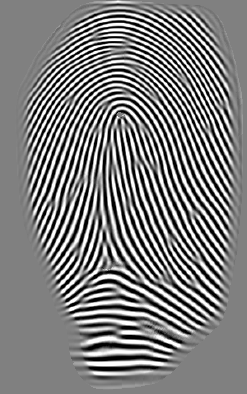}
    \includegraphics[width=0.24\textwidth]{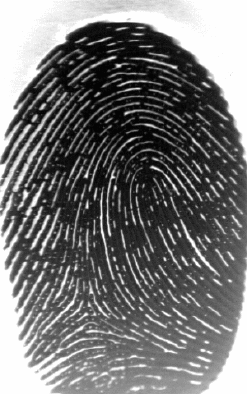}
    \includegraphics[width=0.24\textwidth]{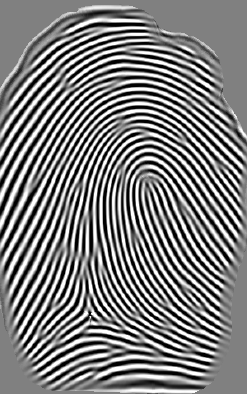}
  \end{center}
  \caption{The enhanced image 1 (second from left) of finger 10 from FVC2004 database 3
          clearly shows the capability of the curved GF to remove scars
          and in the enhanced 
          image 8 (fourth f.l.) of finger 82 from database 1
          falsely conglutinated ridges are separated by the curved GF.
          Filter parameters are $\sigma_{x} = 8.0$, $\sigma_{y} = 8.0$ 
          and the size of the curved regions is $65 \times 65$ pixels.
  \label{figEnhance1}}
\end{figure}

In the case of image enhancement by straight GFs,
\cite{HongWanJain1998} and other authors (e.g. \cite{HandbookFR2009})
use quadratic windows of size $11 \times 11$ pixels 
and choices for the standard deviation of the Gaussian of $\sigma_{x} = \sigma_{y} = 4.0$,
or very similar values. 
We agree with their arguments 
that the parameter selection of $\sigma_{x}$ and $\sigma_{y}$ involves a trade-off 
between an ineffective filter (for small values of $\sigma_{x}$ and $\sigma_{y}$)
and the risk of creating artifacts 
in the enhanced image (for large values of $\sigma_{x}$ and $\sigma_{y}$).
Moreover, the same reasoning holds true for the size of the window.
In analogy to the situation during the RF estimation (see Figure \ref{figComparisonRF}), 
enlarging a rectangular window in a region with curved ridge and valley flow
increases the risk for introducing noise and, as a consequence of this, false structures into the enhanced image.

\begin{figure}
  \begin{center}
    \includegraphics[width=0.3\textwidth]{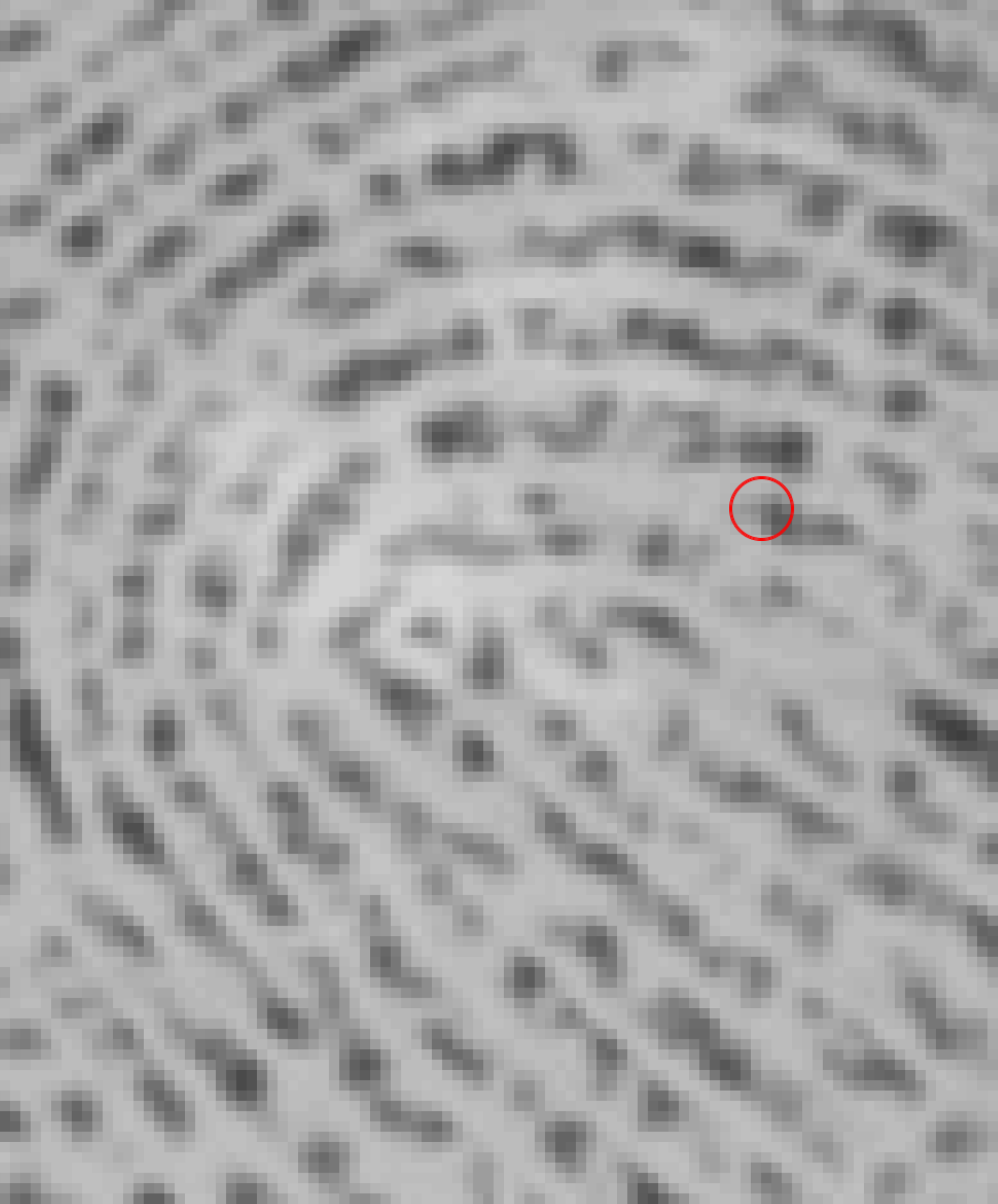}
    \includegraphics[width=0.3\textwidth]{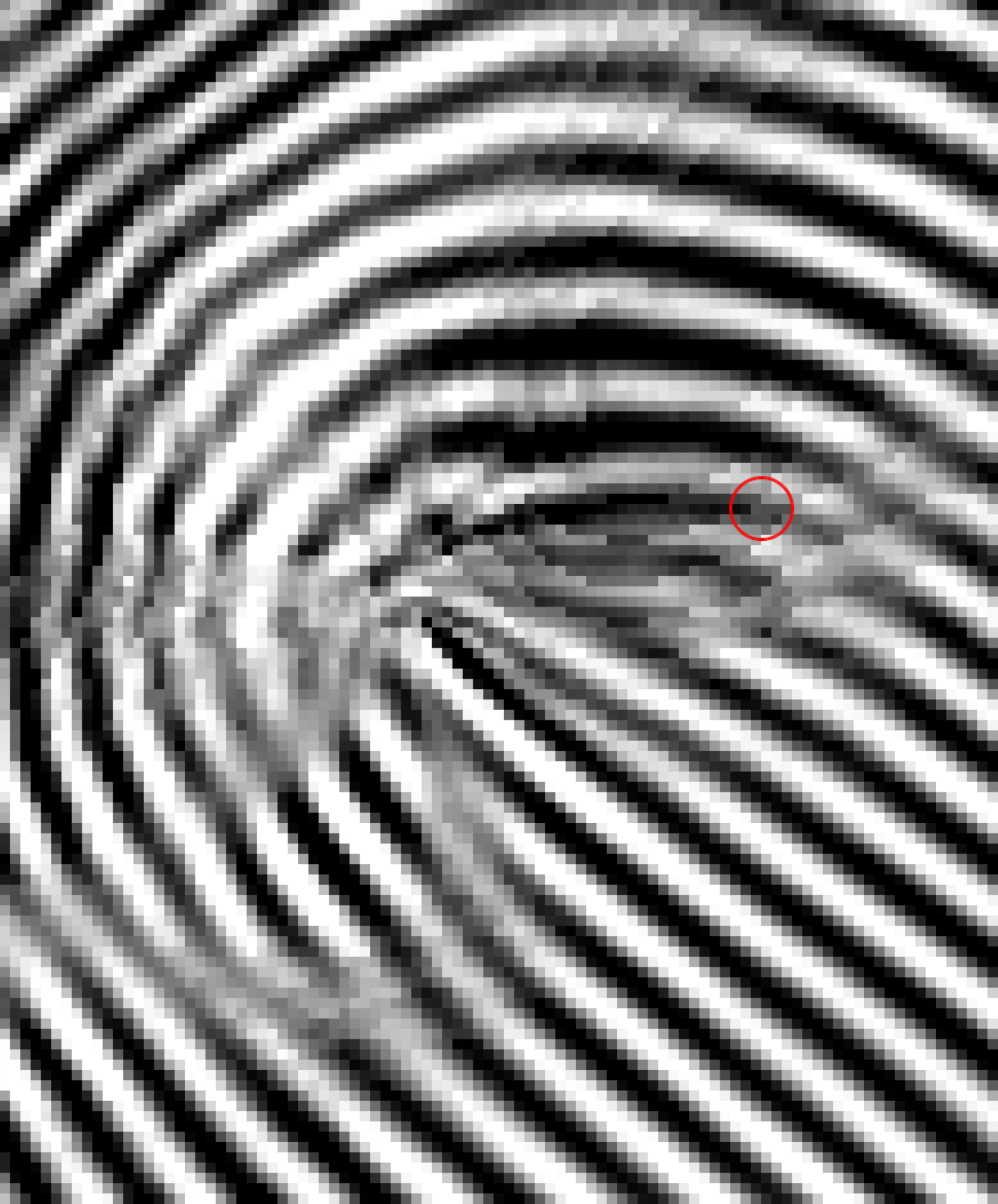}
    \includegraphics[width=0.3\textwidth]{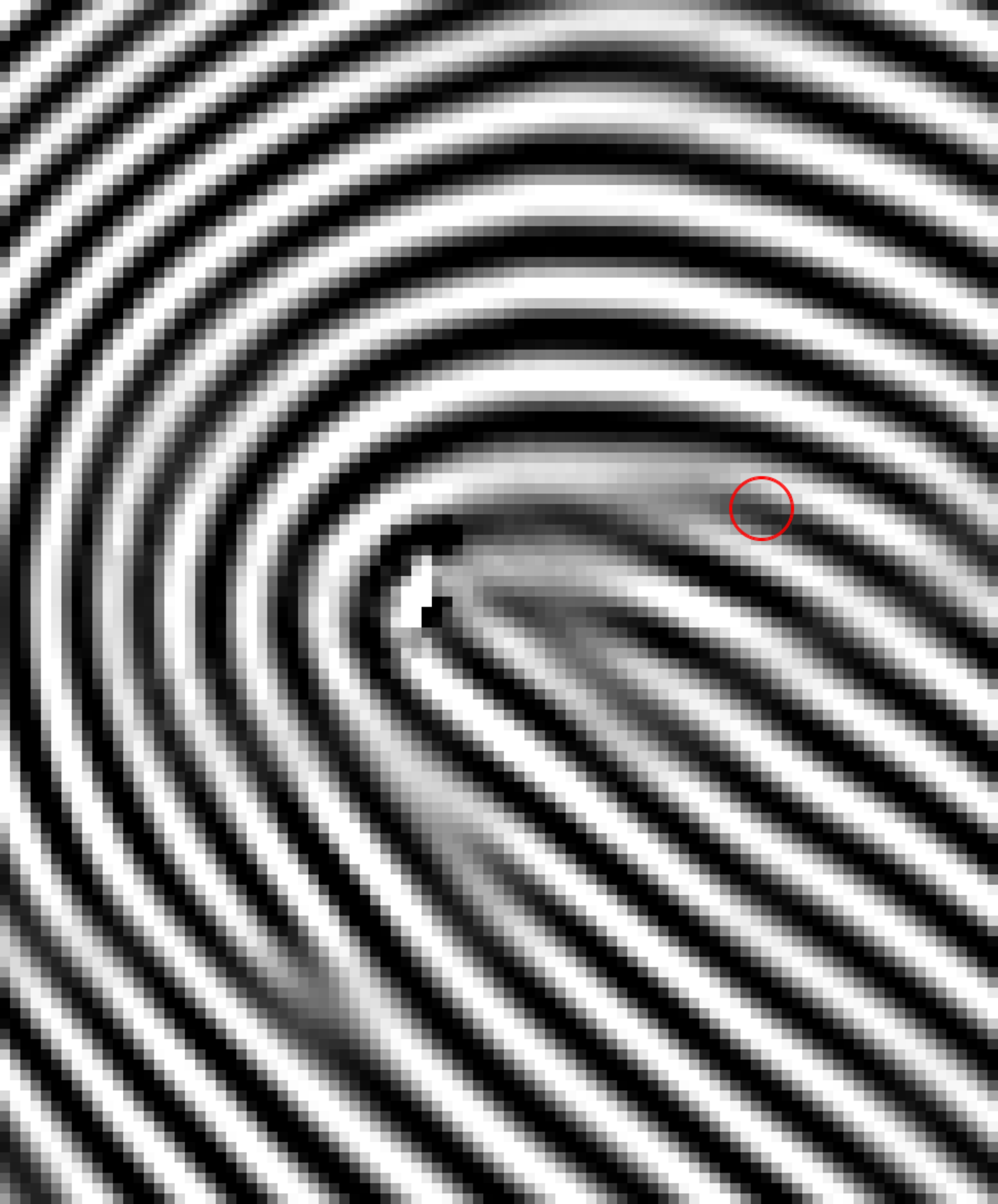}    
  \end{center}
  \caption{The detail on the left (impression 1 of finger 90 in FVC2004 database 4)
           is enhanced by Gabor filtering using rectangular windows (center)
           and curved regions (right). 
           Both filters resort to the same orientation field estimation 
           and the same ridge frequency estimation based on curved regions. 
	   Filter parameters are also identical ($p=16, q=32, \sigma_x=16, \sigma_y=32$),
	   the only difference between the two is the shape of window underlying the Gabor filter.           
	   Artifacts are created by the straight filter which may impair the recognition performance
	   and a true minutia is deleted (highlighted by a red circle).
  \label{figEnhanceDetail}}
\end{figure}

The main advantage of curved Gabor filters is 
that they enable the choice of 
larger curved regions and high values for $\sigma_{x}$ and $\sigma_{y}$
without creating spurious features (see Figures \ref{figEnhance1} and \ref{figEnhanceDetail}).
In this way, curved Gabor filters have a much greater smoothing potential in comparison 
to traditional GF. For curved GFs, the only limitation is the accuracy of the OF and RF estimation,
and no longer the filter itself.

The authors of \cite{ZhuYingZhangHu2006} applied a straight GF for fingerprint enhancement 
and proposed to use a circle instead of a square as the window underlying the GF
in order to reduce the number of artifacts in the enhanced image.
Similarly, we tested an ellipse with major axis $2q + 1$ and minor axis $2p + 1$
instead of the full curved region, i.e. 
in Equation \ref{eqCGF}, only those interpolated gray values of 
array $A_{i,j}$ are considered which are located within the ellipse.
In our tests, both variants achieved similar results on the FVC2004 databases (see Table \ref{tabBZ3}).
As opposed to \cite{ZhuYingZhangHu2006},
the term `circular GF' is used in \cite{ZhangSunGaoCao2008} and \cite{ZhangYang2009}
for denoting the case $\sigma_{x} = \sigma_{y}$.

\section{Results} \label{SecResults}

\subsection{Test Setup}

Two algorithms were employed for matching the original and the enhanced gray-scale images.
The matcher ``BOZORTH3'' is based on the
NIST biometric image software package (NBIS) \cite{NBIS},
applying MINDTCT for minutiae extraction and BOZORTH3 for template matching.
The matcher ``VeriFinger 5.0 Grayscale'' is derived from the Neurotechnology VeriFinger 5.0 SDK.

For the verification tests, we follow the FVC protocol
in order to ensure comparability of the results
with \cite{FronthalerKollreiderBigun2008} and other researchers.
2800 genuine and 4950 impostor recognition attempts were conducted
for each of the FVC databases. 
Equal error rates (EERs) were calculated as described in \cite{FVC2000}.

\subsection{Verification tests}

\begin{table}[ht!]
\begin{center}
\begin{tabular}{|c|c|c|c|c|} \hline 
{Matcher: \textbf{MINDTCT and BOZORTH3}} & \multicolumn{3}{c}{ } &\\ \hline
Enhancement method & DB1 & DB2 & DB3 & DB4 \\ \hline
Original images & 14.5 & 9.5 & 6.2 & 7.3 \\ \hline
Traditional Gabor filter & & & & \\
Hong, Wan, Jain \cite{HongWanJain1998} & \rb{(16.9)} & \rb{14.4} & \rb{7.1} & \rb{9.8} \\ \hline
Short time Fourier transform (STFT) analysis& & & & \\
Chikkerur, Cartwright, Govindaraju \cite{ChikkerurCartwrightGovindaraju2007} & \rb{(19.1)} & \rb{11.9} & \rb{7.6} & \rb{10.9} \\ \hline
Pyramid-based fitering & & & & \\
Fronthaler, Kollreider, Bigun \cite{FronthalerKollreiderBigun2008} & \rb{12.0} & \rb{8.2} & \rb{5.0} & \rb{7.0}  \\ \hline 
{Coherence-enhancing anisotropic diffusion filtering} &  &  &  &   \\ 
{Gottschlich and Sch\"onlieb \cite{GottschlichSchoenlieb2012}} & \rb{10.0} & \rb{6.4} & \rb{5.0} & \rb{6.0} \\ \hline \hline
Straight Gabor filters:&\multicolumn{3}{c}{} &\\\hline 
{x-signature based RF estimation \cite{HongWanJain1998}} &  &  &  &   \\ 
{$11 \times 11$, NN, F, $\sigma_{x} = 4.0$, $\sigma_{y} = 4.0$} & \rb{12.9} & \rb{9.6} & \rb{12.8} & \rb{9.5} \\ \hline
Curved Gabor filters:&\multicolumn{3}{c}{} &\\\hline 
$21 \times 21$, NN, E, $\sigma_{x} = 4.0$, $\sigma_{y} = 4.0$& 10.2 & 6.0 & 4.8 & 6.0 \\ \hline
$33 \times 65$, NN, F, $\sigma_{x} = 4.0$, $\sigma_{y} = 4.0$& 10.5 & 5.9 & 4.8 & 6.5  \\ \hline
$33 \times 65$, NN, E, $\sigma_{x} = 4.0$, $\sigma_{y} = 4.0$& 9.7 & 6.3 & 5.1 & 6.5  \\ \hline
\hline
\multicolumn{1}{|c|}{{Combining curved Gabor filters and}} & \multicolumn{4}{|c|}{} \\ 
\multicolumn{1}{|c|}{{coherence-enhancing anisotropic diffusion filtering \cite{GottschlichSchoenlieb2012}:}} & \multicolumn{4}{|c|}{} \\ \hline
Max rule & 9.0 & 5.0 & 4.2 & 5.4 \\ \hline
Sum rule & 9.3 & 4.8 & 3.6 & 5.2 \\ \hline
Template cross matching & 8.9 & 4.3 & 3.4 & 4.9 \\ \hline \hline
\multicolumn{1}{|c|}{Matcher: \textbf{VeriFinger 5.0 Grayscale}} & \multicolumn{4}{|c|}{} \\ \hline
Original images & 8.8 & 6.8 & 5.3 & 2.0 \\ \hline
Curved Gabor filters:&\multicolumn{3}{c}{} &\\\hline
$21 \times 21$, NN, E, $\sigma_{x} = 4.0$, $\sigma_{y} = 4.0$& 6.4 & 3.5 & 1.7 & 1.3 \\ \hline
$33 \times 65$, NN, F, $\sigma_{x} = 4.0$, $\sigma_{y} = 4.0$& 6.9 & 3.7 & 2.0 & 1.4  \\ \hline
$33 \times 65$, NN, E, $\sigma_{x} = 4.0$, $\sigma_{y} = 4.0$& 6.8 & 3.8 & 1.8 & 1.7  \\ \hline

\end{tabular}
\caption{EERs in \% for matchers Bozorth3 and VeriFinger on the original and enhanced images of FVC2004 \cite{FVC2004}.
         Parentheses indicate that only a small foreground area of the fingerprints was useful for recognition.
         Results listed in the top four rows are cited from \cite{FronthalerKollreiderBigun2008}.
         Parameters of the curved Gabor filters:
         size of the curved region, 
         interpolation method (NN = nearest neighbor), 
         considered pixels (F = full curved region, E = elliptical), 
         standard deviations of Gaussian.
         \label{tabBZ3}}
\end{center}
\end{table}

Curved Gabor filters were applied for enhancing the images of FVC2004 \cite{FVC2004}.
Several choices for $\sigma_{x}$,~$\sigma_{y}$, the size of the curved region 
and interpolation methods were tested.
EERs for some combinations of filter parameters are reported in Table \ref{tabBZ3}.
Other choices for the size of the curved region and the standard deviations of the Gaussian
resulted in similar EERs.
Relating to the interpolation method, only results for nearest neighbor are listed,
because replacing it by bilinear or bicubic interpolation 
did not lead to a noticeable improvement in our tests.

In order to compare the enhancement performance of curved Gabor filters for low quality images
with existing enhancement methods, matcher BOZORTH3 was applied to the enhanced images of FVC2004
which enables the comparison with the traditional GF proposed in \cite{HongWanJain1998},
short time Fourier transform (STFT) analysis \cite{ChikkerurCartwrightGovindaraju2007}
and pyramid-based image filtering \cite{FronthalerKollreiderBigun2008} (see Table \ref{tabBZ3}).

Furthermore, in order to isolate the influence of the OF estimation and segmentation 
on the verification performance, 
we tested the x-signature method \cite{HongWanJain1998} for RF estimation 
and straight Gabor filters in combination with our OF estimation and segmentation.
EERs are listed in the second and sixth row of Table \ref{tabBZ3}. 
In comparison to the results of the cited implementation 
which applied an OF estimation and segmentation as described in \cite{HongWanJain1998},
this led to lower EERs on DB1 and DB2, a higher EER on DB3 and a similar performance on DB4.
In comparison to the performance on the original images, an improvement was observed on the first database
and a deterioration on DB3 and DB4.
Visual inspection of the enhanced images on DB3 showed that the increase of the EER was caused largely
by incorrect RF estimates of the x-signature method.

Moreover, we combined minutiae templates which were extracted by MINDTCT 
from images enhanced by curved Gabor filters 
and from images enhanced by anisotropic diffusion filtering. 
A detailed representation of this combination can be found in \cite{GottschlichSchoenlieb2012}
and results are listed in Table \ref{tabBZ3}.
To the best of our knowledge, this combination performed 
with the lowest EERs on the FVC2004 databases  
which have been achieved so far using MINDTCT and BOZORTH3.

The matcher referred to as VeriFinger 5.0 Grayscale 
has a built-in enhancement step which can not be turned off,
so that the results for the original images in Table \ref{tabBZ3}
are obtained on matching images which were also enhanced 
(by an undisclosed procedure of the commercial software).
Results using this matcher were included in order to show that
even in the face of this built-in enhancement,
the proposed image smoothing by curved Gabor filters leads 
to considerable improvements in verification performance.

\section{Conclusions} \label{SecConclusions}

The present work describes a method for ridge frequency estimation using curved regions
and image enhancement by curved Gabor filters.
For low quality fingerprint images, in comparison to existing enhancement methods
improvements of the matching performance were shown.

Besides matching accuracy, speed is an important factor for fingerprint recognition systems.
Results given in Section \ref{SecResults} were achieved using a proof of concept implementation 
written in Java. 
In a first test of a GPU based implementation on a Nvidia Tesla C2070, 
computing the RF image using curved regions of size $33 \times 65$ pixels 
took about 320~ms and applying curved Gabor filters of size $65 \times 33$ pixels took about 280~ms.
The RF estimation can be further accelerated, if an estimate is computed 
only e.g. for every fourth pixel horizontally and vertically instead of a pixel-wise computation.
These computing times indicate the practicability of the presented method for on-line verification systems.

In our opinion, the potential for further improvements 
of the matching performance rests upon a better OF estimation.
The combined method delineated in Section \ref{SecOF} produces fewer 
erroneous estimations than each of the individual methods, 
but there is still room for improvement.
As long as OF estimation errors occur,
it is necessary to choose the size of the curved Gabor filters
and the standard deviations of the Gaussian envelope with care
in order to balance strong image smoothing while avoiding spurious features.
Future work includes an exploration of a locally adaptive choice of these parameters,
depending on the local image quality, and e.g. the local reliability of the OF estimation.
In addition, it will be of interest to apply the curved region based RF estimation 
and curved Gabor filters to latent fingerprints.

\section*{Acknowledgments}

The author would like to thank Thomas Hotz, Stephan Huckemann, Preda Mih\u{a}ilescu and Axel Munk 
for their valuable comments,
and Daphne B\"ocker for her work on the GPU based implementation.

\end{document}